\documentclass{article}

\usepackage[preprint]{neurips_2025}

\usepackage{tikz}
\usepackage{eso-pic}

\usepackage[utf8]{inputenc} 
\usepackage[T1]{fontenc}    
\usepackage{hyperref}       
\usepackage{url}            
\usepackage{booktabs}       
\usepackage{amsfonts}       
\usepackage{nicefrac}       
\usepackage{microtype}      
\usepackage{xcolor}         
\usepackage{amsmath}
\usepackage{graphicx}
\usepackage{amsfonts}       
\usepackage{multirow}       
\usepackage{adjustbox}
\usepackage{makecell}
\usepackage{float}

\title{The Butterfly Effect in Pathology: Exploring Security in Pathology Foundation Models}

\author{%
\textbf{Jiashuai Liu\textsuperscript{\rm 1}\thanks{Equal contribution.}, 
Yingjia Shang\textsuperscript{\rm 2}\footnotemark[1], 
Yingkang Zhan\textsuperscript{\rm 1}, 
Di Zhang\textsuperscript{\rm 1}, 
Yi Niu\textsuperscript{\rm 1},} \\
\textbf{Dong Wei\textsuperscript{\rm 3}, 
Xian Wu\textsuperscript{\rm 3}, 
Zeyu Gao\textsuperscript{\rm 4}, 
Chen Li\textsuperscript{\rm 1}\thanks{Corresponding author.}, 
Yefeng Zheng\textsuperscript{\rm 5}}\footnotemark[2] \\
\textsuperscript{\rm 1}School of Computer Science and Technology, Xi’an Jiaotong University, Xi’an, China \\
\textsuperscript{\rm 2}Heilongjiang University, Harbin, China \\
\textsuperscript{\rm 3}Tencent Jarvis Lab, Shenzhen, China \\
\textsuperscript{\rm 4}Department of Oncology, University of Cambridge, UK \\
\textsuperscript{\rm 5}Medical Artificial Intelligence Laboratory, Westlake University, Hangzhou, China \\
\texttt{liujs@stu.xjtu.edu.cn} \\
}

\begin{document}

\maketitle

\begin{abstract}
With the widespread adoption of pathology foundation models in both research and clinical decision support systems, exploring their security has become a critical concern. 
However, despite their growing impact, the vulnerability of these models to adversarial attacks remains largely unexplored.
In this work, we present the first systematic investigation into the security of pathology foundation models for whole slide image~(WSI) analysis against adversarial attacks.
Specifically, we introduce the principle of \textit{local perturbation with global impact} and propose a label-free attack framework that operates without requiring access to downstream task labels.
Under this attack framework, we revise four classical white-box attack methods and redefine the perturbation budget based on the characteristics of WSI.
We conduct comprehensive experiments on three representative pathology foundation models across five datasets and six downstream tasks.
Despite modifying only 0.1\% of patches per slide with imperceptible noise, our attack leads to downstream accuracy degradation that can reach up to 20\% in the worst cases. 
Furthermore, we analyze key factors that influence attack success, explore the relationship between patch-level vulnerability and semantic content, and conduct a preliminary investigation into potential defence strategies.
These findings lay the groundwork for future research on the adversarial robustness and reliable deployment of pathology foundation models. Our code is publicly available at: https://github.com/Jiashuai-Liu-hmos/Attack-WSI-pathology-foundation-models\footnote{This work was conducted during Jiashuai Liu and Yingjia Shang’s visit to the Medical Artificial Intelligence Laboratory at Westlake University.}.
\end{abstract}

\section{Introduction}

\AddToShipoutPicture*{%
  \put(80,675){\includegraphics[height=6ex]{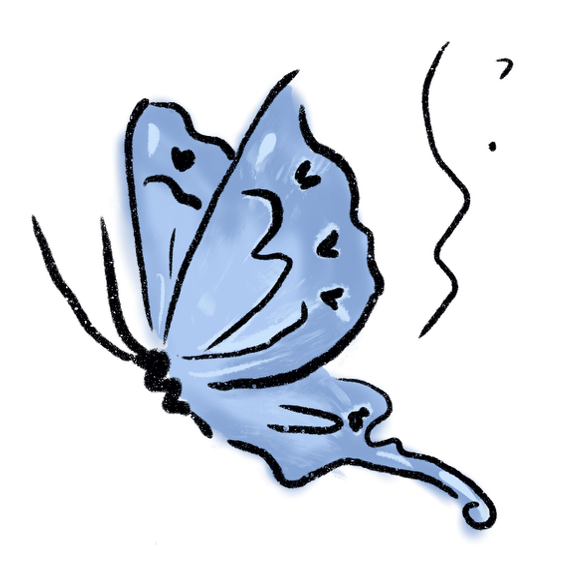}}%
}

Foundation models in computational pathology have emerged as essential tools for a wide range of tasks, including disease diagnosis, prognosis prediction, and biomarker discovery \cite{vaidya2025molecular,xiang2025vision,zhang2025accelerating}. 
Compared to early patch-level foundation models that require task-specific aggregation training \cite{chen2024towards,lu2024visual,vorontsov2024virchow,zimmermann2024virchow2}, recent whole slide image (WSI)-level pathology foundation models (WSI-FMs) integrate pretrained aggregation modules, demonstrating promising performance and superior generalization in downstream few-shot pathology analysis scenarios \cite{xu2024whole,ding2024multimodal}. 
These models leverage WSI-level self-supervised strategies such as masked region reconstruction, intra-slide contrastive learning and cross-modal alignment, enabling robust representation learning under low-resource conditions \cite{ding2024multimodal}. 

Despite their success, the security of these WSI-FMs remains largely underexplored \cite{patch_attack,irmakci2023tissue,thota2024demonstration}. Potential attacks could lead to serious consequences such as misdiagnosis.
Laleh et al. \cite{patch_attack} primarily focus on evaluating the vulnerability of patch-level encoders under labeled settings, while neglecting the complete WSI-level pipeline that includes feature aggregation modules.
In contrast to natural image tasks, the gigapixel scale of WSIs makes conventional adversarial attacks infeasible without substantial adaptation. Furthermore, in real-world clinical settings, a single WSI may be associated with multiple downstream tasks (\textit{e.g.}, classification, grading, and mutation prediction). Yet, adversaries often lack knowledge of the specific task being deployed. 
Consequently, traditional label-dependent attack strategies become impractical in these scenarios.

\begin{figure}[t]
  \centering
  \includegraphics[width=1.0\linewidth]{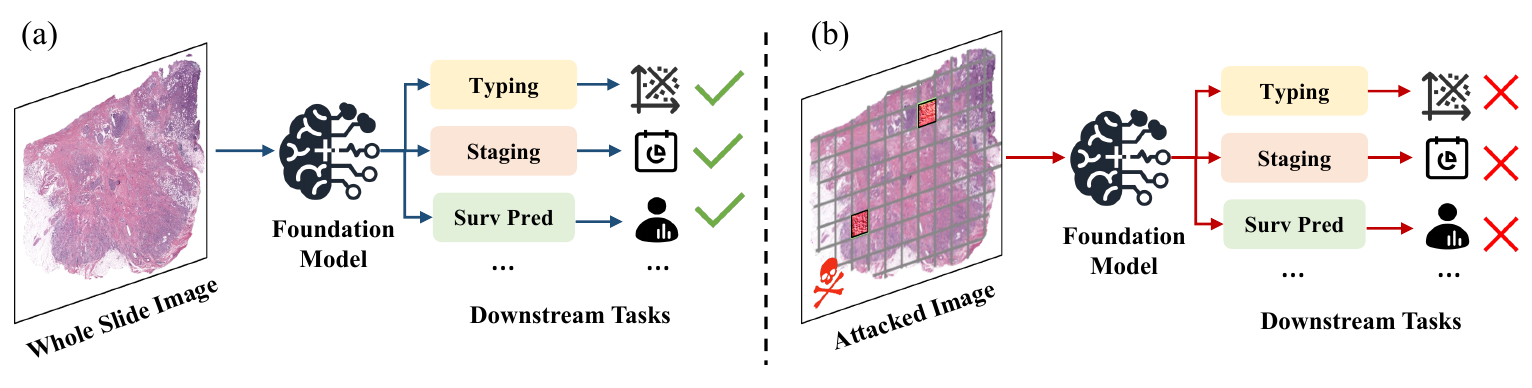}
  \caption{Example of prediction shift under adversarial attack. While the original model predicts correctly on the clean input (a), the adversarially perturbed version (b) leads to incorrect predictions across various downstream tasks.}
  \label{fig:pipline}
\end{figure}

To address these challenges, this work presents the first systematic investigation into the security of WSI-FMs under adversarial attacks. 
Reflecting real-world clinical deployment scenarios, we define a practical attack setting in which model parameters are accessible due to their availability during deployment.
Under this constraint, we develop a label-free adversarial attack framework tailored for pathology foundation models, as shown in Figure \ref{fig:pipline}.
Considering the gigapixel scale of WSIs, we introduce the principle of local perturbation with global impact, where only a small subset of selected patches is perturbed to disrupt the global slide-level representation, thereby degrading the model’s performance on downstream diagnostic tasks. We support this framework by tailoring four well-established white-box attack techniques to operate effectively in a label-free setting.
We redefine the perturbation budget to balance perturbation strength and patch-level coverage, aiming to ensure the attack’s stealth and practicality.
We conduct comprehensive experiments across five widely used pathology datasets and six representative downstream tasks, evaluating the security of three leading WSI-FMs: CHIEF \cite{chief}, PRISM \cite{shaikovski2024prism}, and TITAN \cite{ding2024multimodal}. We further conduct empirical investigations into the key factors contributing to adversarial attack success.
Finally, we explore a lightweight defence strategy that leverages the  natural robustness of WSI-FMs against random noise. 
This approach provides initial insights into improving the adversarial security of pathology models and supports future research in trustworthy computational pathology.

\paragraph{Contributions.} 
This paper makes the following key contributions:  
(1) We propose the first label-free adversarial attack framework explicitly designed for WSI-FMs, introducing the principle of local perturbation with global impact.  
(2) We establish a systematic security benchmark across five datasets and six diagnostic tasks to evaluate the adversarial robustness of three representative models, i.e., CHIEF, PRISM, and TITAN.  
(3) We find that CHIEF demonstrates the strongest adversarial robustness. However, it still suffers a substantial degradation in performance. The number of selected patches and attack strategy play a more decisive role than perturbation magnitude in attack effectiveness.
(4) We analyze critical factors influencing attack success and propose a lightweight defence method leveraging the intrinsic resilience of foundation models to random noise.

\section{Related work}
\subsection{Foundation models for pathology}

In recent years, the introduction of foundation models has significantly enhanced the generalization ability of computational pathology across diverse tasks and multi-center settings \cite{liang2024wsi,lu2023foundational,vaidya2025molecular,xiang2025vision,xu2024multimodal,zhang2025accelerating}. 
Through self-supervised pretraining on large-scale pathology datasets, traditional patch-level models have shown strong capabilities in capturing local cytological and morphological patterns, forming the basis for advances in computational pathology \cite{chen2024towards,lu2024visual,vorontsov2024virchow,zimmermann2024virchow2}.
However, as the need for patient-level and WSI-level modeling has grown, research has increasingly shifted toward WSI-FMs, which take the entire slide as the modeling unit. These models produce holistic slide-level representations and have demonstrated strong performance in tasks such as survival analysis, molecular prediction, and tumor subtyping \cite{ding2024multimodal,shaikovski2024prism,xu2024whole,zhang2025accelerating}.

Representative models include CHIEF \cite{chief}, PRISM \cite{shaikovski2024prism}, and TITAN \cite{ding2024multimodal}, all of which adopt a two-stage architecture consisting of feature extraction and aggregation modules. 
CHIEF employs image-text contrastive learning to align visual features with both molecular labels and textual descriptions, thereby enabling effective cross-modal representation learning. In contrast, PRISM extends its applicability to broader downstream tasks by leveraging large-scale alignment with clinical reports, even beyond the scope of explicitly annotated data. 
TITAN enhances global feature representations by modeling inter-patch relationships and adopting a staggered training strategy that aligns WSIs with associated captions and pathology narratives.
Despite their strong generalization, the adversarial robustness of these models remains underexplored, particularly under imperceptible perturbations—raising concerns about their reliability in clinical deployment.

\subsection{Adversarial attack}

Adversarial attacks have revealed significant vulnerabilities in the security and robustness of deep neural networks (DNNs) \cite{advbenchmark,advreview,attackguided,introcnn}. Adversarial attacks are typically categorized into white-box and black-box settings, depending on the level of access the adversary has to the internal components of the model. 
In white-box attacks, the adversary is assumed to have full access to the model's architecture and gradient information. In contrast, black-box attacks restrict the adversary to only input-output interactions, without access to model parameters or gradients.

This study focuses on a deployment-aligned adversarial attack setting, where the feature extraction and aggregation modules at the WSI-level are accessible in a white-box manner, while the downstream task model remains unknown and inaccessible (black-box). This hybrid scenario, which combines white-box access to the WSI-level backbone with a black-box downstream model, makes adversarial methods that leverage white-box information particularly applicable and effective.

In conventional classification tasks, adversarial attack methods perturb the input to induce misclassification. The \textit{Fast Gradient Sign Method} (FGSM) \cite{fgsm} was one of the earliest proposed white-box approaches. It performs a single-step update in the gradient direction to maximize the loss with respect to the true label. To enhance the attack strength, the \textit{Basic Iterative Method} (BIM) \cite{bim} extends FGSM into multiple small-step iterations, increasing perturbation effectiveness while clipping at each step to keep the perturbation bounded. 
Building upon this, the \textit{Momentum Iterative Method} (MIM) \cite{mim} introduces momentum accumulation to stabilize the update direction. In contrast, the \textit{Carlini and Wagner Attack} (C\&W) \cite{cw} formulates the attack as a constrained optimization problem, minimizing perturbation magnitude while ensuring attack success.

Despite the extensive study of adversarial attacks in the domain of natural images \cite{pgd,adversarialtrain}, their application in digital pathology remains nascent. 
Gigapixel resolution and heterogeneity in pathology images pose significant challenges to traditional adversarial attack methods.
While a few studies have investigated patch-level adversarial attacks in pathology \cite{patch_attack}, WSI-level models remain underexplored. 
Yet these models better reflect real-world use scenarios with integrated patch feature extraction and aggregation modules, thus deserving a systematic evaluation of their adversarial vulnerabilities.

\section{Threat model}

To simulate a more realistic deployment scenario, we propose a practically grounded adversarial threat model. In clinical workflows, WSIs are typically processed by a pre-trained pathology foundation model to extract representations. These representations are subsequently reused across multiple downstream diagnostic tasks. The shared usage of WSI-level features makes the foundation model a critical and vulnerable component within the pipeline.

A WSI-level pathology foundation model usually consists of a patch-level feature extractor $\phi(\cdot)$ and a feature aggregation network $\mathcal{A}(\cdot)$, both of which are independently pre-trained and support offline inference and gradient access. During the processing pipeline, each WSI is divided into a set of image patches denoted as $\left\{x_i \in \mathbb{R}^{H \times W \times C} \right\}_{i=1}^N$, where $N$ is the number of patches. Each patch is encoded by the feature extractor into a feature vector $f_i = \phi(x_i) \in \mathbb{R}^d$, forming the feature set $\mathcal{F} = \left\{f_1, f_2, \dots, f_N\right\}$. The feature aggregation network $\mathcal{A}(\cdot)$ then generates a WSI-level representation: $z = \mathcal{A}(\mathcal{F}) \in \mathbb{R}^{d'}$. 
In practice, the feature extractor and aggregator of the given WSI-FM are typically used in a pair.

Specifically, we assume that the adversary has full access to the architecture and parameters of the foundation model, including both the feature extractor and the feature aggregator. In contrast, the downstream task-specific modules are assumed to be inaccessible. The adversary has no knowledge of their structure, training objectives, label definitions, or prediction outputs. Under this setting, the adversary does not aim to mislead the prediction of any specific downstream task. Instead, the objective is to modify the WSI-level features extracted by the foundation model, thereby indirectly affecting the performance of all downstream tasks that rely on these features.

\begin{figure}[t]
  \centering
  \includegraphics[width=1.0\linewidth]{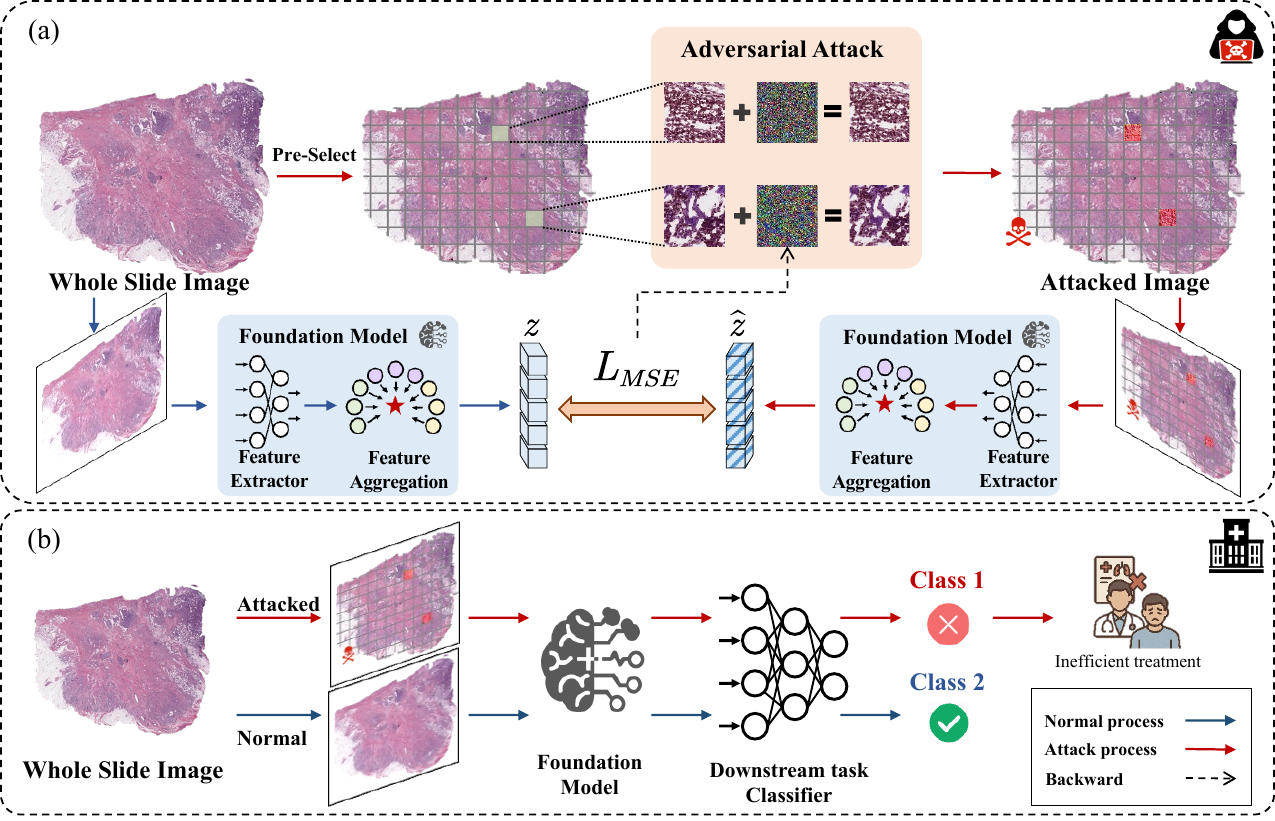}
  \caption{Overview of the proposed adversarial attack framework and its clinical attack scenario.
(a) A label-free adversarial attack framework targeting WSI-FMs. Localized perturbations are applied to selected patches, leading to shifts in global representations. (b) The adversarial perturbation leads to incorrect predictions, introducing clinical risks that may result in misdiagnosis and improper treatment decisions.}
  \label{fig:architecture}
\end{figure}

\section{Adversarial attack on pathology foundation models}
\subsection{Local perturbation with global impact}

Considering the gigapixel nature of WSIs, we design a ``local perturbation–global impact'' attack mechanism to explore how subtle local perturbations can be amplified through the WSI feature aggregation process as shown in Figure \ref{fig:architecture}. Specifically, $K$ patches are selected from the total $N$ patches for attack, where $K \ll N$. The attack set is denoted as $\mathcal{I},\ |\mathcal{I}|=K$. Unlike traditional attacks, this method only perturbs selected patches while keeping all other patches in the image unchanged. 
For each patch $x_i$, a perturbation $\delta_i$ is added if it is in $\mathcal{I}$:
\begin{equation}
\widehat{x}_i = 
\begin{cases}
x_i + \delta_i, & \text{if } x_i \in \mathcal{I}, \\ 
x_i, & \text{otherwise,}
\end{cases}
\end{equation}
where $|\delta_i|_\infty \leq \epsilon$ indicates that each pixel can be perturbed by at most $\epsilon$. 
The perturbed WSI is then encoded using the same feature extractor, resulting in $\widehat{f}_i = \phi(\widehat{x}_i) \in \mathbb{R}^d$, and the full set of features after perturbation is denoted as $\widehat{\mathcal{F}} = \left\{\widehat{f}_1, \widehat{f}_2, \dots, \widehat{f}_N\right\}$.
Finally, the aggregated WSI-level representation after attack is obtained via the feature aggregator as $\widehat{z} = \mathcal{A}(\widehat{\mathcal{F}})$.

To enable adversarial attacks without access to downstream task labels and predictions, we reformulate the attack objective as maximizing the discrepancy between the original WSI representation $z$ and its perturbed counterpart $\widehat{z}$:
\begin{equation}
   \mathcal{I},\, \delta = \arg\max_{\mathcal{I},\, \delta} \; \text{MSE}\left(\widehat{z}, z\right),
\end{equation}
where $\mathcal{I}$ denotes the subset of patches selected for attack, and $\delta$ represents the perturbations applied to these patches. 
Hence, we define the loss function as $L_{\text{MSE}} = \text{MSE}\left(\widehat{z}, z\right)$. 
By indirectly maximizing the loss function, we drive the features away from their original decision boundaries, thereby affecting the performance of unseen downstream tasks. Subsequent experiments further validate the effectiveness of this loss function design.

\subsection{Label-free adaptation of white-box attack methods}
\label{attack_object}

We implement the proposed framework by adapting four classic white-box adversarial attack methods—FGSM, BIM, MIM, and C\&W \cite{fgsm,bim,mim,cw}.
Specifically, we adapt the attack objectives of these methods to the label-free setting of WSI-FMs by maximizing the discrepancy between the original and perturbed representations in the feature space, without relying on ground-truth labels. In addition, unlike traditional attacks that modify the entire input, our method updates only the selected patches via gradient-based optimization, leaving all others unchanged. Taking \textbf{FGSM} for example, the patch update process is as follows:
\begin{equation}
\widehat{x}_i = x_i + \epsilon \cdot \text{sign}(\nabla_{x_i} L_{\text{MSE}}).
\label{eq:local_fgsm}
\end{equation}

Furthermore, we extend this objective to the other three mainstream attack methods.
\textbf{BIM} is a multi-step version of FGSM. It performs updates with a small step size at each iteration and constrains perturbations within a bounded region. Its feature-driven formulation is:
\begin{equation}
\widehat{x}_{i,t+1} = \text{Clip}_{x_p, \epsilon} \left( \widehat{x}_{i,t} + \alpha \cdot \text{sign} \left( \nabla_{\widehat{x}_{i,t}} L_{\text{MSE}} \right) \right),
\label{eq:bim}
\end{equation}
where $\alpha$ is the step size and $\text{Clip}_{x_p, \epsilon}(\cdot)$ projects the update back into the $\epsilon$-ball neighborhood of the selected patch.

\textbf{MIM} builds upon BIM by incorporating a momentum term to stabilize the direction of the updates:
\begin{equation}
g_{t+1} = \mu \cdot g_t + \frac{\nabla_{\widehat{x}_{i,t}} L_{\text{MSE}}}{\| \nabla_{\widehat{x}_{i,t}} L_{\text{MSE}} \|_1},
\label{eq:mim_momentum}
\end{equation}
\begin{equation}
\widehat{x}_{i,t} = \text{Clip}_{x_i, \epsilon} \left( \widehat{x}_{i,t} + \alpha \cdot \text{sign}(g_{t+1}) \right),
\label{eq:mim_update}
\end{equation}
where $\mu$ is the momentum coefficient and $g_t$ is the accumulated gradient.

\textbf{C\&W attack} is reformulated as a joint optimization problem to minimize perturbation magnitude while maximizing feature deviation. The objective is defined as:
\begin{equation}
\min_{\delta} \left\{ \| \widehat{x}_i - x_i \|_2 - c \cdot L_{\text{MSE}} \right\},
\label{eq:cw}
\end{equation}
where $c$ is a weighting coefficient that balances the trade-off between perturbation size and feature discrepancy. To ensure consistency across experimental conditions, we apply the same projection operation to adversarial examples generated by the C\&W attack as used in BIM. After optimization, we project the final adversarial example back into the $L_\infty$ ball centered at the original input with radius $\epsilon$, thereby standardizing perturbation bounds across different methods and ensuring fair and comparable results.

\subsection{Perbution budget}
Unlike natural images that typically have fixed and reasonable input dimensions, WSIs contain tens of thousands of patches and can reach gigapixel resolutions. Attacking all patches simultaneously is not only computationally prohibitive but also compromises the stealth of the attack. Therefore, we extend the standard adversarial budget definition from natural images by introducing two orthogonal constraints: (1) the \textbf{perturbation scope budget}, defined as the number of attacked patches $K$, and (2) the \textbf{perturbation magnitude budget}, denoted by $\epsilon$, which limits the per-pixel change within each selected patch.

\subsection{Attack strategy}
To ensure diversity in patch selection, we randomly sample $K$ patches from the WSI for attack.
Based on the $K$ selected patches, we propose two attack strategies to model different computational constraints in practice: (1) \textbf{Sequential attack}, which emulates adversarial behavior under limited computational resources. 
In each iteration, a perturbation is applied to a single patch and is updated on top of the previously accumulated perturbation. The WSI-level representation is recomputed after each step, and the perturbation is optimized to maximize the representation shift.  (2) \textbf{Parallel attack}, which perturbs multiple patches simultaneously in each iteration, jointly optimizing their perturbations. It reflects high-intensity attack scenarios where parallel computation is feasible.

\section{Experiments}

\subsection{Settings}

In this study, we employ the CLAM toolbox \cite{lu2021data} to segment tissue regions from WSIs and extract image patches of size 512×512 pixels at 20× magnification. We select six downstream tasks across five widely used histopathology datasets, all of which have previously demonstrated strong performance using WSI-FMs. 
Specifically, the tasks include: estrogen receptor (ER) status classification (binary) on TCGA-BRCA \cite{brca}; renal cancer subtype classification (three-class) on TCGA-RCC \cite{davis2014somatic,cancer2016comprehensive,cancer2013comprehensive}; lung cancer subtype classification (binary) on TCGA-LUNG \cite{cancer2012comprehensive,cancer2014comprehensive}; tumor detection (binary) on Camelyon16 \cite{litjens20181399}; and both IDH mutation status prediction (binary) and coarse-grained brain tumor subtyping (twelve-class, following the setting of UNI \cite{chen2024towards}) on EBRAINS \cite{appukuttan2023ebrains}.
Detailed information about the datasets is provided in Appendix.
To perform the downstream tasks, we train and evaluate multilayer perceptron (MLP) classifiers under the five-fold cross-validation setting for all the datasets.

Considering the high experimental time cost and varying dataset sizes, we randomly select 300 WSIs of each task for attack to ensure data representativeness, as well as a timely and consistent evaluation of each task.
It is worth noting that, for each selected WSI, we evaluate the impact of adversarial perturbations using the MLP classifier whose training data does not include the specific WSI, in order to avoid leakage.
We benchmark three state-of-the-art WSI-FMs: CHIEF, PRISM, and TITAN. Following each model's standard pipeline, we extract patch-level features, aggregate them into slide-level representations.
Finally, we input these features into an MLP for downstream classification. We evaluate attack effectiveness using two metrics: Accuracy Drop (AD) and Attack Success Rate (ASR), defined in \cite{wu2021performance}. 
For both metrics, higher values indicate greater degradation in model performance caused by adversarial attacks.

\renewcommand{\arraystretch}{1.5}

\begin{table}[htbp]
\centering
\caption{Performance of three pathology WSI-FMs on six downstream tasks under attack by various methods.
Classification accuracy (\%) is reported for the `No Attack' setting, whereas both accuracy and Accuracy Drop (AD, \%) are reported for each attack method. Max AD values of each foundation model on each dataset are bloded.}
\label{tab1}
\begin{adjustbox}{width=\textwidth}
\begin{tabular}{cccccccc}
\toprule
                          & Dataset & TCGA-BRCA     & TCGA-RCC       & TCGA-LUNG      & Camelyon        & \multicolumn{2}{c}{EBrains}        \\ \cline{2-8}\rule{0pt}{5pt} 
                          & Task    & \makecell{ER\\(2-class)}  & \makecell{Type\\(3-class)} & \makecell{Type\\(2-class)} & \makecell{Tumor\\(2-class)} & \makecell{Subtype\\(12-class)} & \makecell{IDH\\(2-class)} \\ \cline{2-8} 
\multirow{3}{*}{No Attack}   &CHIEF & 87.63 & 96.39 & 93.14 & 84.31 & 90.58 & 91.05 \\
&PRISM & 83.28 & 89.95 & 93.06 & 95.30 & 89.90 & 88.63 \\
&TITAN & 85.62 & 94.50 & 96.18 & 98.32 & 95.89 & 91.97 \\ 
\midrule

\multirow{3}{*}{C\&W}  & CHIEF & 87.37\ (\(\downarrow\)0.26) & 95.10\ (\(\downarrow\)1.29) & 92.88\ (\(\downarrow\)0.26) & 76.59\ (\(\downarrow\)7.72) & 87.67\ (\(\downarrow\)2.91) & 90.55\ (\(\downarrow\)0.50) \\
 & PRISM & 76.59\ (\(\downarrow\)6.69) & 74.57\ (\(\downarrow\)15.38) & 80.47\ (\(\downarrow\)12.59) & 84.14\ (\(\downarrow\)11.16) & 77.57\ (\(\downarrow\)12.33) & 83.53\ (\(\downarrow\)5.10) \\
 & TITAN & 79.52\ (\pmb{\(\downarrow\)6.10}) & 81.62\ (\(\downarrow\)12.88) & 82.73\ (\(\downarrow\)13.45) & 73.57\ (\(\downarrow\)24.75) & 80.31\ (\(\downarrow\)15.58) & 86.29\ (\(\downarrow\)5.68) \\
\midrule

\multirow{3}{*}{FGSM}     & CHIEF & 87.46\ (\(\downarrow\)0.17) & 93.64\ (\(\downarrow\)2.75) & 92.27\ (\(\downarrow\)0.87) & 51.34\ (\pmb{\(\downarrow\)32.97}) & 86.04\ (\(\downarrow\)4.54) & 89.63\ (\pmb{\(\downarrow\)1.42)} \\
 & PRISM & 57.11\ (\pmb{\(\downarrow\)26.17}) & 44.42\ (\pmb{\(\downarrow\)45.53}) & 54.51\ (\pmb{\(\downarrow\)38.55}) & 45.39\ (\pmb{\(\downarrow\)49.91}) & 28.25\ (\pmb{\(\downarrow\)61.65}) & 51.51\ (\pmb{\(\downarrow\)37.12}) \\
 & TITAN & 80.18\ (\(\downarrow\)5.44) & 79.12\ (\pmb{\(\downarrow\)15.38}) & 85.33\ (\(\downarrow\)10.85) & 41.78\ (\pmb{\(\downarrow\)56.54}) & 66.61\ (\pmb{\(\downarrow\)29.28}) & 84.45\ (\(\downarrow\)7.52) \\

\midrule
\multirow{3}{*}{MIM}      & CHIEF & 87.29\ (\pmb{\(\downarrow\)0.34}) & 93.30\ (\(\downarrow\)3.09) & 91.75\ (\(\downarrow\)1.39) & 60.32\ (\(\downarrow\)23.99) & 86.73\ (\(\downarrow\)3.85) & 90.22\ (\(\downarrow\)0.83) \\
 & PRISM & 69.31\ (\(\downarrow\)13.97) & 54.04\ (\(\downarrow\)35.91) & 69.36\ (\(\downarrow\)23.70) & 59.31\ (\(\downarrow\)35.99) & 49.32\ (\(\downarrow\)40.58) & 64.80\ (\(\downarrow\)23.83) \\
 & TITAN & 80.60\ (\(\downarrow\)5.02) & 80.24\ (\(\downarrow\)14.26) & 80.90\ (\pmb{\(\downarrow\)15.28}) & 42.28\ (\(\downarrow\)56.04) & 69.01\ (\(\downarrow\)26.88) & 82.78\ (\pmb{\(\downarrow\)9.19}) \\
\bottomrule

\multirow{3}{*}{BIM}       & CHIEF & 87.37\ (\(\downarrow\)0.26) & 92.78\ (\pmb{\(\downarrow\)3.61}) & 91.58\ (\pmb{\(\downarrow\)1.56)} & 55.79\ (\(\downarrow\)28.52) & 85.79\ (\pmb{\(\downarrow\)4.79}) & 89.97\ (\(\downarrow\)1.08) \\
 & PRISM & 63.21\ (\(\downarrow\)20.07) & 45.88\ (\(\downarrow\)44.07) & 62.24\ (\(\downarrow\)30.82) & 47.82\ (\(\downarrow\)47.48) & 35.27\ (\(\downarrow\)54.63) & 54.52\ (\(\downarrow\)34.11) \\
 & TITAN & 80.77\ (\(\downarrow\)4.85) & 83.68\ (\(\downarrow\)10.82) & 85.94\ (\(\downarrow\)10.24) & 43.62\ (\(\downarrow\)54.70) & 76.28\ (\(\downarrow\)19.61) & 85.20\ (\(\downarrow\)6.77) \\
\bottomrule
\end{tabular}
\end{adjustbox}

\end{table}

\subsection{Vulnerability of WSI-FMs to attack}
\label{main_exp}
We first investigate the vulnerability of WSI-FMs in the extreme case of $K=1$, i.e., only a single patch---about 0.1\% of patches per slide---is perturbed.
Specifically, for each WSI, we randomly select a patch to perturb and set the perturbation budget $\epsilon$ to 4/255, which is generally considered imperceptible. 
Additional experimental details are provided in Appendix.
We repeat the single-random-patch attack four times and report the average result. 
Experimental results in Table \ref{tab1} show that applying imperceptible perturbations to fewer than 0.1\% of patches can still result in a substantial performance drop of up to 20\% in downstream tasks, such as cancer detection on the Camelyon16 dataset. 
The standard deviations of the repeated experiments are provided in Appendix.
This demonstrates a significant security threat to WSI-FMs. Among the evaluated models, CHIEF exhibits the strongest adversarial robustness. We will discuss the potential underlying reasons in Section \ref{butterfly}

\begin{figure}[t]
  \centering
  \includegraphics[width=1.0\linewidth]{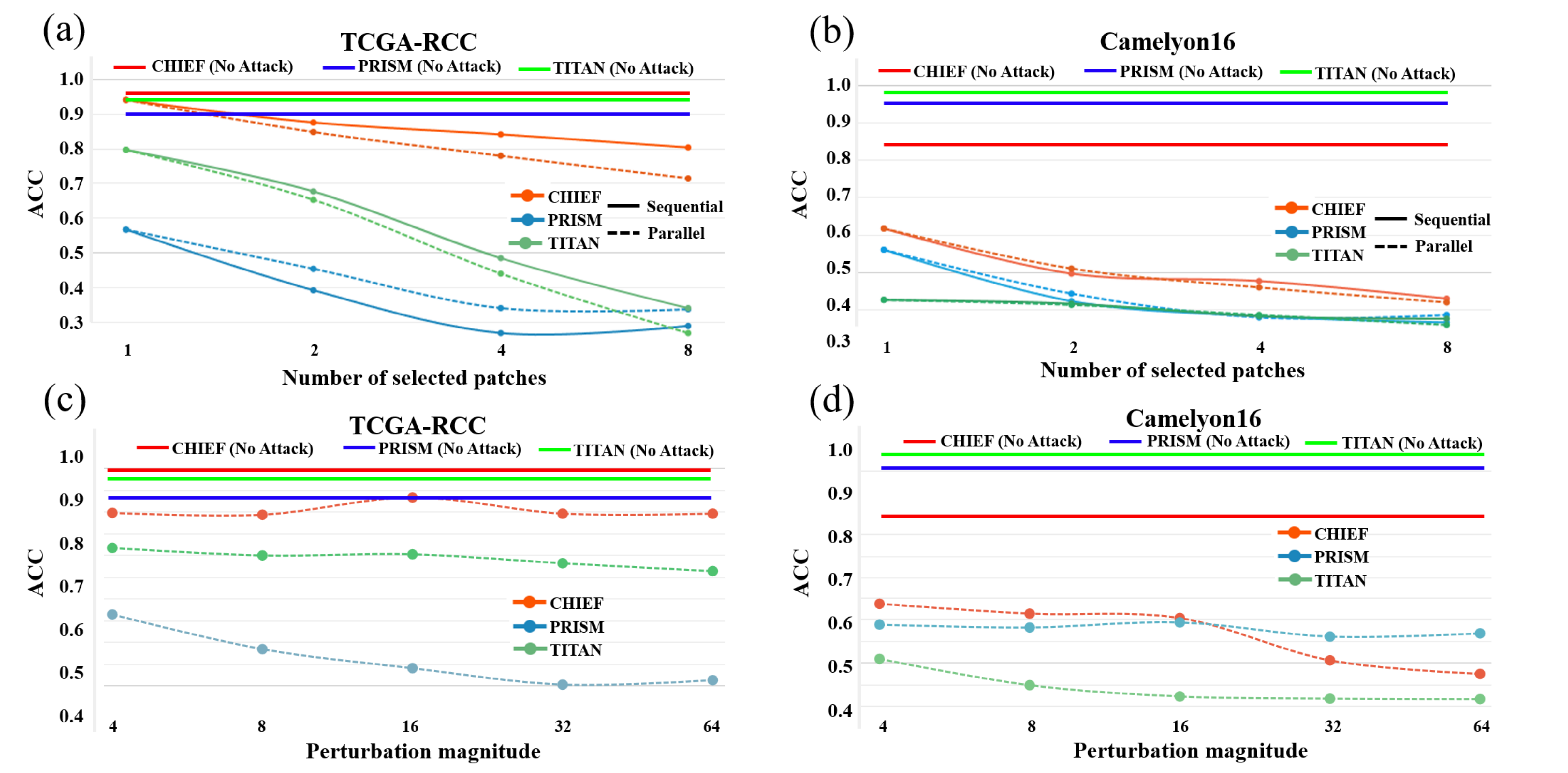}
  \caption{Evaluation of attack performance under varying perturbation budgets on two downstream tasks (TCGA-RCC and Camelyon16).
(a–b) Accuracy comparison with different numbers of perturbed patches using parallel and sequential attack strategies.
(c–d) Accuracy comparison with different perturbation magnitudes (i.e., $\epsilon$). Solid and dashed lines represent the sequential and parallel attack strategies, respectively. Horizontal lines indicate classification accuracy without attack.}
  \label{fig:exp23}
\end{figure}

\subsection{Perturbation budget and attack strategy analysis}
\label{budget}
Building on our earlier definition, we investigate how the perturbation budget, including both scope and magnitude, influences attack effectiveness. 
For perturbation scope, we increase the number of perturbed patches from one to eight and compare the two attack strategies: parallel and sequential attacks. 
The former perturbs multiple patches jointly in a single optimization step, while the latter iteratively updates each patch in sequence. 
We adopt the MIM model and evaluate its performance on two tasks: RCC subtype classification and Camelyon16 tumor detection. 
The results are shown in Figure \ref{fig:exp23}(a–b). 
As the number of perturbed patches increases, both strategies' attack effectiveness steadily improves. 
However, we observe no clear performance advantage for either parallel or sequential attack.
In practice, if the computing resources are limited, Sequential Attack could be employed.
Otherwise, parallel attack could be used for better time efficiency.

In addition, we examine the impact of perturbation magnitude by varying $\epsilon$ across {4, 8, 16, 32, and 64}, under the single-patch attack setting (Section \ref{main_exp}). 
The results are presented in Figure \ref{fig:exp23}(c-d). As the perturbation magnitude increases, the attack becomes progressively more effective. However, this improvement is less pronounced compared to the impact of the perturbation scope. 
Excessive perturbation on a single patch can also compromise visual integrity, making it perceptible to humans.
Therefore, single-patch attacks are less effective than multi-patch ones.

\begin{figure}[t]
  \centering
  \includegraphics[width=0.9\linewidth]{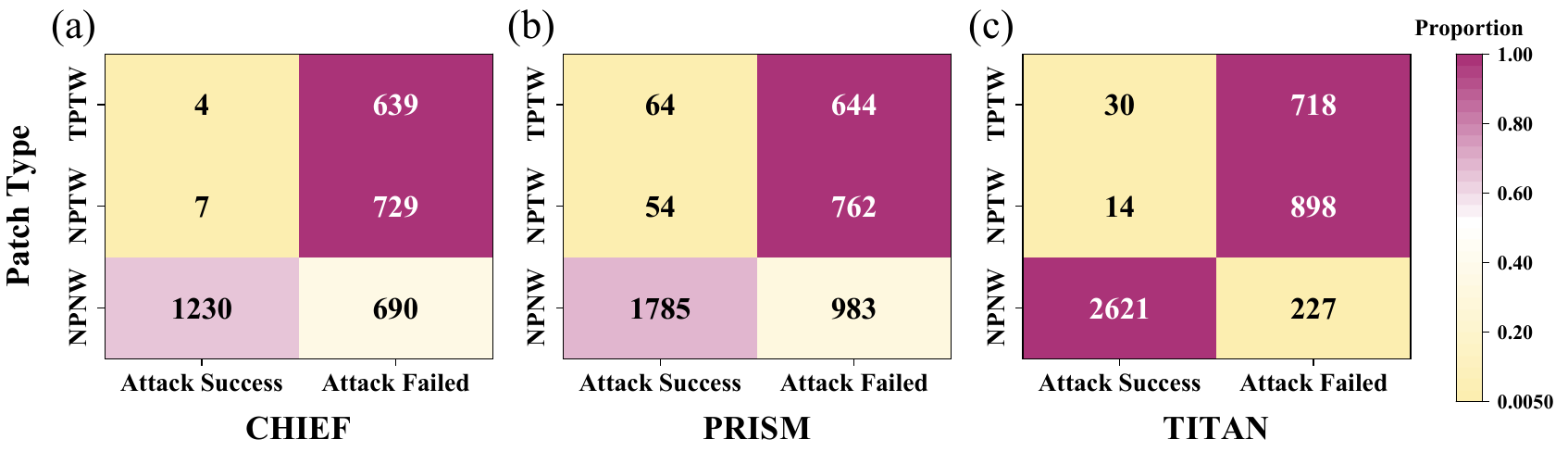}
  \caption{Distribution of attack outcomes across different patch types.
TPTW: tumor patch from tumor WSI; NPTW: normal patch from tumor WSI; NPNW: normal patch from normal WSI.}
  \label{fig:exp4}
\end{figure}

\subsection{Which patches are more vulnerable?}

WSIs in pathology exhibit a high degree of histological complexity and spatial heterogeneity. In recent years, Multiple Instance Learning (MIL) methods have made significant progress under weak supervision \cite{gao2024accurate,ilse2018attention,liu2024pamil,lu2021data,shao2021transmil}. By leveraging class-guided mechanisms, these models can automatically attend to diagnostically relevant regions and capture patch-level semantic information without precise annotations. However, there remains a lack of investigation into whether the semantic properties of individual patches are associated with their vulnerability to adversarial attacks.

To explore this, we conduct an experiment on the Camelyon16 dataset. This dataset provides pixel-level annotations that distinguish cancerous from non-cancerous regions, allowing each patch to be explicitly labeled as either tumor or normal.
For each WSI that is correctly predicted originally, we randomly select 16 patches and perform single-patch attacks on them individually.
To investigate the relationship between semantic class and adversarial vulnerability, we compute the attack success rate separately for tumor and normal patches.
As shown in Figure \ref{fig:exp4}, the attacks targeting normal WSIs are more likely to succeed. This is consistent with the MIL assumption, where the presence of even a single tumor patch can influence the slide-level prediction.
In contrast, for tumor WSIs, the attack success rate remains relatively low regardless of whether the perturbed patch is tumor or normal. Interestingly, and somewhat counterintuitively, perturbing normal patches within tumor WSIs can also confuse the model to output a normal prediction.
These findings suggest a complex relationship between adversarial vulnerability and the semantic class of individual patches.


\begin{figure}[t]
  \centering
  \includegraphics[width=0.9\linewidth]{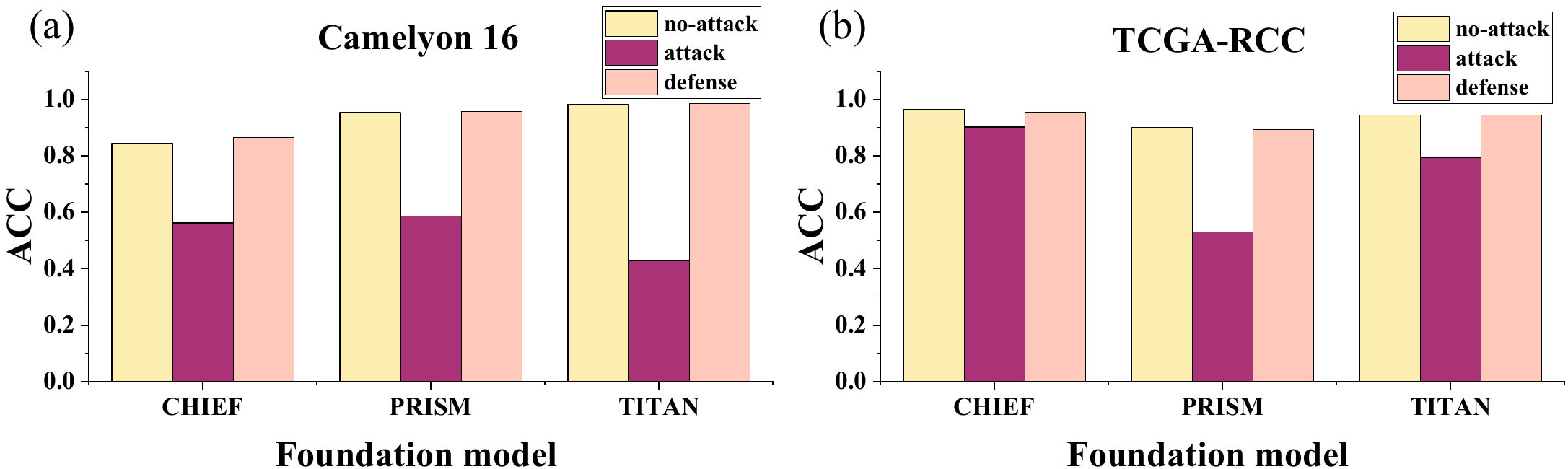}
  \caption{Defence performance on two downstream tasks, reporting the prediction accuracy of each foundation model under no attack, attack, and defence.}
  \label{fig:exp5}
\end{figure}

\subsection{Defence strategy}
\label{defence}

Adversarial attacks significantly degrade the performance of WSI-FMs across various tasks, posing potential risks in clinical applications. Despite their generalization ability and moderate robustness, these models remain vulnerable to carefully designed perturbations, highlighting the need for effective defence strategies. Observing that self-supervised pretraining imparts a degree of resilience to random noise, we propose a lightweight defence mechanism that injects small-scale uniform noise (±1/255) into all input patches. This approach aims to disrupt the structured, gradient-aligned nature of adversarial perturbations.
As shown in Figure \ref{fig:exp5}, the experiments on Camelyon16 and TCGA-RCC datasets indicate that this simple strategy surprisingly reduces the effectiveness of adversarial attacks and maintains the models' original performance, suggesting its potential as a practical and efficient defence solution with further development.

\section{Discussion}
\subsection{Butterfly effect in pathology}
\label{butterfly}
From the above experiments, we observe an interesting phenomenon: the adversarial robustness of the three WSI-FMs follows the order CHIEF > TITAN > PRISM. Notably, this ordering does not align with their performance under non-adversarial conditions. Further analysis reveals a positive correlation between attack success rate and model size, as measured by the number of parameters (see Appendix for details).
We hypothesize that the network can represent more complex piecewise-linear decision boundaries as model capacity increases. In such settings, small input perturbations are more likely to be locally amplified near the decision boundary. Moreover, larger models tend to exhibit sharper gradients in both direction and magnitude, making them inherently more sensitive to subtle changes in input space.
This observation echoes the notion of the \textbf{``butterfly effect''} in deep learning: minute, seemingly imperceptible perturbations can trigger disproportionate changes in prediction outcomes due to the high complexity and nonlinearity of large-scale models, thereby exposing significant security vulnerabilities of WSI-FMs.

\subsection{Limitations and broader impact}
\label{limitation}

While this study provides a systematic analysis of adversarial vulnerabilities in WSI-level pathology foundation models, it has several limitations as an initial exploration. 
First, the attack success is shown to be highly dependent on individual patches' cancerous status, yet this study only evaluates random selection without incorporating structural or attention-based priors.
We expect to develop more effective attack strategies along this line in the future.
Second, the current loss function is limited to MSE. 
The effectiveness of alternative objectives remains to be investigated. 
Third, the study does not disentangle the respective vulnerabilities of the feature extractor and the aggregator module, which would offer more profound insights into where adversarial weaknesses reside.

As foundation models become increasingly integrated into computational pathology, ensuring their secure and trustworthy deployment is of growing importance. 
On the one hand, this work provides a first-step investigation into the adversarial robustness of WSI-FMs, offering both a systematic evaluation of existing models and initial insights for future research on the security of pathology foundation models. Our proposed defence strategy also sheds light on potential directions for developing more robust systems. On the other hand, given that our attack method can significantly degrade model performance with imperceptible perturbations, we emphasize the need to remain vigilant against potential misuse and the risks such attacks may pose to the clinical deployment of such models.

\section{Conclusion}
In this work, we presented the first systematic study on the adversarial security of WSI-FMs. We proposed a practical label-free attack framework where model parameters are accessible but task-specific labels remain unknown. By introducing the principle of local perturbation with global impact and redefining the perturbation budget, we adapted classical white-box attack methods to the pathology domain. Through extensive experiments on five datasets and six downstream tasks, we demonstrated that even state-of-the-art models such as CHIEF can experience significant performance degradation under carefully crafted adversarial perturbations. 
Our findings revealed a strong correlation between attack success and perturbation scope, rather than perturbation magnitude alone. 
Finally, we showed that adding small-scale random noise is a lightweight yet effective defence strategy, offering initial insights toward enhancing the robustness and trustworthiness of computational pathology models in safety-critical applications.

\bibliographystyle{plain}
\bibliography{ref}

\newpage
\appendix
\section{Dataset descriptions}

We evaluate the foundation models across six downstream tasks drawn from five widely used histopathology datasets. To approximate real-world deployment settings, all classifiers are trained using five-fold cross-validation on each dataset's complete set of WSIs. Below, we summarize the class distributions for each task.
\begin{itemize}
    \item TCGA-BRCA (ER status, binary): 779 slides (179 Negative, 600 Positive).
    \item TCGA-RCC (renal cancer subtype, 3-class): 918 slides (90 KICH, 537 KIRC, 291 KIRP).
    \item TCGA-LUNG (lung cancer subtype, binary): 1,018 slides (514 LUAD, 504 LUSC).
    \item Camelyon16 (tumor detection, binary): 399 slides (239 Normal, 160 Tumor-positive).
    \item EBRAINS (IDH mutation status, binary): 873 slides (540 Negative, 333 Positive).
    \item EBRAINS (coarse-grained brain tumor subtyping, 12-class): 2,318 slides following the setting of UNI \cite{chen2024towards}, with class counts as follows — adult-type diffuse gliomas (837), meningiomas (430), mesenchymal non-meningothelial tumors (190), sellar region tumors (184), circumscribed astrocytic gliomas (173), ependymal tumors (96), hematolymphoid tumors (90), glioneuronal and neuronal tumors (88), cranial and paraspinal nerve tumors (81), pediatric-type diffuse low-grade gliomas (70), metastatic tumors (47), and embryonal tumors (32).
\end{itemize}

\section{Overview of WSI-FMs}

In Section \ref{butterfly}, we analyze the performance of three WSI-FMs under adversarial attacks. The results reveal a robustness ranking of CHIEF > TITAN > PRISM. Interestingly, this ordering is inversely correlated with the number of model parameters, as shown in Table \ref{tab:FM-summary}. 
We hypothesize that as the number of parameters increases, the models become more vulnerable to small gradient-based perturbations. When well aligned with the loss landscape, these perturbations are more likely to be locally amplified near the decision boundary. Consequently, larger models exhibit heightened sensitivity to even minor perturbations in the input space.
This observation echoes the \textbf{``butterfly effect''} referenced in our title, highlighting not only the vulnerability of pathology foundation models to adversarial attacks, but also the chaotic behavior that may emerge from the black-box nature of large-scale models.

\begin{table}[htbp]
\centering
\caption{Pretraining strategies, parameter counts, and robustness ranks for feature aggregation modules and feature extractors constituting each WSI-FM.}
\label{tab:FM-summary}
\begin{adjustbox}{width=\textwidth}
\begin{tabular}{ccccccc}
\toprule
\makecell{\textbf{Feature} \\ \textbf{Aggregation}} 
& \makecell{\textbf{Pretrain} \\ \textbf{Method}} 
& \makecell{\textbf{Params} \\ (M)} 
& \makecell{\textbf{Feature} \\ \textbf{Extractor}} 
& \makecell{\textbf{Pretrain} \\ \textbf{Method}} 
& \makecell{\textbf{Params} \\ (M)} 
& \makecell{\textbf{Robustness} \\ \textbf{Rank}} \\
\midrule
CHIEF & Weakly Supervised & 1.19  & CTransPath  & SRCL    & 27.52  & 1st \\
TITAN & iBOT+COCA         & 158.87 & CONCH v1.5  & SSL     & 306.11 & 2nd \\
PRISM & COCA              & 557.70 & Virchow     & DINOv2  & 631.23 & 3rd \\
\bottomrule
\end{tabular}
\end{adjustbox}
\end{table}

\section{Experimental settings and hardware configuration}

Unless otherwise specified, we adopt a consistent set of hyperparameters for all adversarial attacks. Specifically, BIM, MIM, and C\&W attacks are each run for 20 iterations. The step size for BIM and MIM is set to $\alpha = 1$, while the loss balancing coefficient for C\&W is set to $c = 1$. The perturbation budget is fixed at $\epsilon = 4/255$, and all attacks are conducted under the single-patch setting. In experiments investigating the effect of perturbation magnitude, we adjust the step size to $\alpha = \epsilon / 4$ to ensure the perturbation reaches the specified budget within a limited number of iterations.

To ensure computational feasibility, we cap the number of patches per WSI at 7,000. For slides containing more than 7,000 patches, we extract a contiguous subset of 7,000 patches from the original sequence. This constraint is applied consistently across all experiments, including the no-attack baseline. Empirically, we observe that this restriction does not compromise downstream task performance.

Our experiments were conducted on a server equipped with eight NVIDIA RTX 4090 GPUs and another server with eight NVIDIA A100 GPUs. 
To optimize memory usage during experiments, different foundation models were assigned to different GPU types based on their memory requirements.
For example, on the TCGA-RCC subtype classification task discussed in Section 5.2, the MIM attack on CHIEF takes approximately 0.54 hours on a 4090 GPU. Attacks on PRISM and TITAN require around 1.75 hours (4090) and 9.33 hours (A100), respectively.
In particular, defence experiments require repeated feature extraction, which significantly increases the runtime. On TCGA-RCC, the complete attack-defence process for CHIEF takes approximately 10 hours on a 4090 GPU, 12 hours for PRISM (also on a 4090), and up to 60 hours for TITAN on an A100 GPU.
It is worth mentioning that although we used multiple GPUs of different types to optimize experiment efficiency, all experiments can be executed on a single A100 GPU.

\section{Additional quantitative results}

To ensure the completeness of our experimental results, we provide additional quantitative data and corresponding standard deviations in this appendix. Table \ref{tab:main_acc} extends the results from Section \ref{main_exp} by reporting the standard deviations. Table \ref{tab:main_asr} introduces Attack Success Rate (ASR) as an additional evaluation metric for more comprehensive assessment of model robustness. For the perturbation scope experiments in Section \ref{budget}, detailed results and standard deviations are included in Tables \ref{tab:patches_c16} and \ref{tab:patches_rcc}. The experiments exploring perturbation magnitude, also in Section \ref{budget}, are supplemented with exact values and standard deviations in Tables \ref{tab:mag_c16} and \ref{tab:mag_rcc}. Finally, Table \ref{tab:attack-defence} presents the full results of the attack-and-defence experiments discussed in Section \ref{defence}, providing a more complete basis for comparative analysis.

\begin{table}[htbp]
\centering
\caption{Classification accuracy (\%) of three WSI-FMs on six downstream tasks under attack by various methods.
Format: mean $\pm$ standard deviation.
}
\label{tab:main_acc}
\begin{adjustbox}{width=\textwidth}
\begin{tabular}{lccccccc}
\toprule
 & Dataset & TCGA-BRCA & TCGA-RCC & TCGA-LUNG & Camelyon & \multicolumn{2}{c}{EBrains} \\
\cmidrule(lr){2-8}
 & Task & ER (2-class) & Type (3-class) & Type (2-class) & Tumor (2-class) & Subtype (12-class) & IDH (2-class) \\
\midrule
\multirow{3}{*}{No Attack} 
& TITAN & 85.62 & 94.50 & 96.18 & 98.32 & 95.89 & 91.97 \\
& CHIEF & 87.63 & 96.39 & 93.14 & 84.31 & 90.58 & 91.05 \\
& PRISM & 83.28 & 89.95 & 93.06 & 95.30 & 89.90 & 88.63 \\
\midrule
\multirow{3}{*}{C\&W}
& TITAN & $79.52 \pm 1.51$ & $81.62 \pm 0.20$ & $82.73 \pm 1.37$ & $73.57 \pm 1.87$ & $80.31 \pm 1.06$ & $86.29 \pm 0.47$ \\
& CHIEF & $87.37 \pm 0.50$ & $95.10 \pm 0.71$ & $92.88 \pm 1.08$ & $76.59 \pm 0.88$ & $87.67 \pm 1.22$ & $90.55 \pm 0.57$ \\
& PRISM & $76.59 \pm 2.17$ & $74.57 \pm 2.62$ & $80.47 \pm 1.22$ & $84.14 \pm 1.61$ & $77.57 \pm 1.06$ & $83.53 \pm 1.26$ \\
\midrule
\multirow{3}{*}{FGSM}
& TITAN & $80.18 \pm 2.24$ & $79.12 \pm 2.62$ & $85.33 \pm 2.65$ & $41.78 \pm 0.64$ & $66.61 \pm 1.44$ & $84.45 \pm 1.48$ \\
& CHIEF & $87.46 \pm 0.84$ & $93.64 \pm 0.91$ & $92.27 \pm 0.77$ & $51.34 \pm 1.57$ & $86.04 \pm 1.29$ & $89.63 \pm 0.27$ \\
& PRISM & $57.11 \pm 2.68$ & $44.42 \pm 1.80$ & $54.51 \pm 2.52$ & $45.39 \pm 0.96$ & $28.25 \pm 4.12$ & $51.51 \pm 3.31$ \\
\midrule
\multirow{3}{*}{MIM}
& TITAN & $80.60 \pm 0.55$ & $80.24 \pm 1.74$ & $80.90 \pm 2.46$ & $42.28 \pm 0.91$ & $69.01 \pm 1.80$ & $82.78 \pm 1.04$ \\
& CHIEF & $87.29 \pm 0.77$ & $93.30 \pm 1.07$ & $91.75 \pm 0.59$ & $60.32 \pm 2.72$ & $86.73 \pm 1.23$ & $90.22 \pm 0.74$ \\
& PRISM & $69.31 \pm 3.40$ & $54.04 \pm 1.95$ & $69.36 \pm 2.44$ & $59.31 \pm 3.50$ & $49.32 \pm 2.02$ & $64.80 \pm 2.09$ \\
\midrule
\multirow{3}{*}{PGD}
& TITAN & $80.77 \pm 1.04$ & $83.68 \pm 1.60$ & $85.94 \pm 2.51$ & $43.62 \pm 0.87$ & $76.28 \pm 1.32$ & $85.20 \pm 1.86$ \\
& CHIEF & $87.37 \pm 0.92$ & $92.78 \pm 0.49$ & $91.58 \pm 0.17$ & $55.79 \pm 2.75$ & $85.79 \pm 1.30$ & $89.97 \pm 0.47$ \\
& PRISM & $63.21 \pm 2.49$ & $45.88 \pm 1.27$ & $62.24 \pm 1.00$ & $47.82 \pm 1.36$ & $35.27 \pm 1.88$ & $54.52 \pm 3.08$ \\
\bottomrule
\end{tabular}
\end{adjustbox}
\end{table}

\begin{table}[htbp]
\centering
\caption{ASR (\%) of three WSI-FMs on six downstream tasks under attack by various methods.
Format: mean $\pm$ standard deviation.}
\label{tab:main_asr}
\begin{adjustbox}{width=\textwidth}
\begin{tabular}{lccccccc}
\toprule
 & Dataset & TCGA-BRCA & TCGA-RCC & TCGA-LUNG & Camelyon & \multicolumn{2}{c}{EBrains} \\
\cmidrule(lr){2-8}
 & Task & ER (2-class) & Type (3-class) & Type (2-class) & Tumor (2-class) & Subtype (12-class) & IDH (2-class) \\
\midrule
\multirow{3}{*}{C\&W}
& TITAN & $11.52 \pm 5.58$ & $15.73 \pm 7.24$ & $15.61 \pm 7.82$ & $25.68 \pm 11.82$ & $17.59 \pm 6.80$ & $9.00 \pm 3.94$ \\
& CHIEF & $0.29 \pm 0.72$ & $1.44 \pm 0.72$ & $0.75 \pm 0.71$ & $10.96 \pm 1.54$ & $3.02 \pm 0.99$ & $0.94 \pm 0.42$ \\
& PRISM & $10.21 \pm 1.61$ & $18.82 \pm 1.86$ & $14.26 \pm 0.85$ & $12.46 \pm 1.07$ & $13.97 \pm 0.82$ & $7.62 \pm 1.12$ \\
\midrule
\multirow{3}{*}{FGSM}
& TITAN & $10.74 \pm 4.57$ & $18.27 \pm 7.56$ & $13.36 \pm 6.25$ & $58.79 \pm 25.57$ & $31.96 \pm 13.45$ & $11.64 \pm 5.04$ \\
& CHIEF & $0.19 \pm 0.91$ & $3.16 \pm 0.60$ & $1.41 \pm 0.61$ & $42.65 \pm 1.64$ & $4.13 \pm 0.99$ & $1.89 \pm 0.26$ \\
& PRISM & $38.54 \pm 3.43$ & $55.49 \pm 1.80$ & $44.30 \pm 2.87$ & $56.00 \pm 0.98$ & $69.53 \pm 4.05$ & $48.44 \pm 2.93$ \\
\midrule
\multirow{3}{*}{MIM}
& TITAN & $10.55 \pm 4.06$ & $17.27 \pm 7.72$ & $17.96 \pm 8.75$ & $58.19 \pm 25.12$ & $29.02 \pm 13.30$ & $14.09 \pm 5.83$ \\
& CHIEF & $0.38 \pm 0.87$ & $3.61 \pm 1.04$ & $1.69 \pm 0.56$ & $32.35 \pm 2.39$ & $3.23 \pm 1.57$ & $1.42 \pm 0.83$ \\
& PRISM & $20.52 \pm 3.12$ & $42.84 \pm 1.92$ & $28.04 \pm 3.20$ & $40.32 \pm 3.13$ & $45.45 \pm 2.12$ & $31.35 \pm 2.50$ \\
\midrule
\multirow{3}{*}{PGD}
& TITAN & $9.77 \pm 3.89$ & $13.45 \pm 6.46$ & $12.91 \pm 5.94$ & $56.83 \pm 24.53$ & $21.96 \pm 10.36$ & $10.45 \pm 4.88$ \\
& CHIEF & $0.29 \pm 0.95$ & $3.97 \pm 0.44$ & $2.16 \pm 0.31$ & $37.39 \pm 2.14$ & $4.23 \pm 1.26$ & $1.60 \pm 0.31$ \\
& PRISM & $28.96 \pm 2.98$ & $52.55 \pm 1.25$ & $35.46 \pm 1.52$ & $53.41 \pm 1.12$ & $61.34 \pm 2.06$ & $44.82 \pm 3.27$ \\
\bottomrule
\end{tabular}
\end{adjustbox}
\end{table}

\begin{table}[htbp]
\centering
\caption{Classification accuracy (\%) under varying numbers of attacked patches (1, 2, 4, 8) using sequential and parallel strategies on Camelyon16 tumor detection.
Format: mean $\pm$ standard deviation.}
\label{tab:patches_c16}
\begin{adjustbox}{width=\textwidth}
\begin{tabular}{ccccccc}
\toprule
Strategy   & Model & No Attack & 1 & 2 & 4 & 8 \\
\midrule
Sequential & CHIEF & 84.31 & $61.74 \pm 0.19$ & $49.66 \pm 0.39$ & $47.65 \pm 1.04$ & $42.95 \pm 1.48$ \\
Sequential & PRISM & 95.30 & $56.04 \pm 0.70$ & $42.28 \pm 1.89$ & $38.26 \pm 1.43$ & $36.58 \pm 0.02$ \\
Sequential & TITAN & 98.32 & $42.62 \pm 0.19$ & $41.61 \pm 0.86$ & $38.26 \pm 0.13$ & $37.58 \pm 0.04$ \\
\midrule
Parallel   & CHIEF & 84.31 & $61.74 \pm 0.19$ & $51.01 \pm 0.33$ & $45.97 \pm 0.19$ & $41.95 \pm 0.84$ \\
Parallel   & PRISM & 95.30 & $56.04 \pm 0.70$ & $44.30 \pm 0.84$ & $37.92 \pm 0.67$ & $38.59 \pm 1.46$ \\
Parallel   & TITAN & 98.32 & $42.62 \pm 0.19$ & $41.28 \pm 0.19$ & $38.59 \pm 0.34$ & $35.91 \pm 2.18$ \\
\bottomrule
\end{tabular}
\end{adjustbox}
\end{table}

\begin{table}[htbp]
\centering
\caption{Classification accuracy (\%) under varying numbers of attacked patches (1, 2, 4, 8) using sequential and parallel strategies on TCGA-RCC subtyping.
Format: mean $\pm$ standard deviation.}
\label{tab:patches_rcc}
\begin{adjustbox}{width=\textwidth}
\begin{tabular}{llccccc}
\toprule
Strategy   & Model & No Attack & 1 & 2  & 4  & 8  \\
\midrule
Sequential & CHIEF & 96.39 & $94.16 \pm 0.19$ & $87.63 \pm 0.60$ & $84.19 \pm 1.00$ & $80.41 \pm 0.71$ \\
Sequential & PRISM & 89.95 & $56.70 \pm 0.40$ & $39.18 \pm 2.50$ & $26.80 \pm 1.95$ & $28.87 \pm 2.03$ \\
Sequential & TITAN & 94.50 & $79.73 \pm 0.33$ & $67.70 \pm 1.67$ & $48.45 \pm 0.95$ & $34.02 \pm 1.58$ \\
\midrule
Parallel   & CHIEF & 96.39 & $94.16 \pm 0.19$ & $84.88 \pm 0.33$ & $78.01 \pm 0.77$ & $71.48 \pm 1.39$ \\
Parallel   & PRISM & 89.95 & $56.70 \pm 0.40$ & $45.36 \pm 2.34$ & $34.02 \pm 1.91$ & $33.68 \pm 1.73$ \\
Parallel   & TITAN & 94.50 & $79.73 \pm 0.33$ & $65.29 \pm 1.30$ & $43.99 \pm 1.50$ & $26.80 \pm 1.95$ \\
\bottomrule
\end{tabular}
\end{adjustbox}
\end{table}

\newpage
\begin{table}[H]
\centering
\caption{Classification accuracy (\%) under increasing perturbation magnitude ($\epsilon$ from $4/255$ to $64/255$) on Camelyon16 tumor detection.
Format: mean $\pm$ standard deviation.}
\label{tab:mag_c16}
\begin{adjustbox}{width=\textwidth}
\begin{tabular}{lcccccc}
\toprule
Model & No Attack & 4 & 8 & 16 & 32 & 64 \\
\midrule
CHIEF & 84.31 & $63.59 \pm 1.51$ & $61.33 \pm 2.83$ & $60.32 \pm 2.72$ & $50.59 \pm 1.38$ & $47.48 \pm 0.64$ \\
PRISM & 95.30 & $58.81 \pm 1.07$ & $58.14 \pm 3.02$ & $59.31 \pm 3.50$ & $56.04 \pm 1.19$ & $56.80 \pm 2.16$ \\
TITAN & 98.32 & $50.92 \pm 2.36$ & $44.88 \pm 1.51$ & $42.28 \pm 0.91$ & $41.78 \pm 1.41$ & $41.69 \pm 0.74$ \\
\bottomrule
\end{tabular}
\end{adjustbox}
\end{table}

\begin{table}[H]
\centering
\caption{Classification accuracy (\%) under increasing perturbation magnitude ($\epsilon$ from $4/255$ to $64/255$) on TCGA-RCC subtyping.
Format: mean $\pm$ standard deviation.}
\label{tab:mag_rcc}
\begin{adjustbox}{width=\textwidth}
\begin{tabular}{lcccccc}
\toprule
Model & No Attack & 4 & 8  & 16  & 32  & 64  \\
\midrule
CHIEF & 96.39 & $89.78 \pm 0.90$ & $89.35 \pm 1.19$ & $93.30 \pm 1.07$ & $89.60 \pm 1.20$ & $89.60 \pm 1.10$ \\
PRISM & 89.95 & $66.41 \pm 1.80$ & $58.42 \pm 2.82$ & $54.04 \pm 1.95$ & $50.26 \pm 1.72$ & $51.29 \pm 4.28$ \\
TITAN & 94.50 & $81.70 \pm 1.33$ & $79.98 \pm 2.20$ & $80.24 \pm 1.74$ & $78.18 \pm 3.14$ & $76.37 \pm 1.99$ \\
\bottomrule
\end{tabular}
\end{adjustbox}
\end{table}

\begin{table}[H]
\centering
\caption{Classification accuracy (\%) on Camelyon16 tumor detection and TCGA-RCC subtyping under adversarial attack and defence.
Format: mean $\pm$ standard deviation.}
\label{tab:attack-defence}
\begin{adjustbox}{width=\textwidth}
\begin{tabular}{lcccccc}
\toprule
& \multicolumn{3}{c}{Camelyon16} & \multicolumn{3}{c}{TCGA-RCC} \\
\cmidrule(lr){2-4} \cmidrule(lr){5-7}
Method & PRISM & CHIEF & TITAN & PRISM & CHIEF & TITAN \\
\midrule
No Attack & 95.30 & 84.31 & 98.32 & 89.95 & 96.39 & 94.50 \\
Attack    & $58.56 \pm 2.27$ & $56.21 \pm 1.24$ & $42.79 \pm 1.49$ & $52.92 \pm 2.85$ & $90.21 \pm 0.20$ & $79.30 \pm 2.09$ \\
defence   & $95.64 \pm 0.61$ & $86.58 \pm 1.06$ & $98.49 \pm 0.19$ & $89.35 \pm 0.28$ & $95.45 \pm 0.33$ & $94.33 \pm 0.44$ \\
\bottomrule
\end{tabular}
\end{adjustbox}
\end{table}

\end{document}